\pdfoutput=1
\documentclass[10pt]{article}
\usepackage{paperonecol}
\usepackage{latexsym}
\usepackage[utf8]{inputenc}
\usepackage{microtype}
\usepackage{amsmath,amssymb,booktabs,tabularx,array}
\usepackage{longtable}
\usepackage{graphicx}
\usepackage{placeins}
\usepackage{colortbl}
\definecolor{tablegray}{gray}{0.92}
\usepackage{enumitem}
\usepackage{url}
\usepackage[colorlinks=true,citecolor=blue,linkcolor=black,urlcolor=blue]{hyperref}
\newcolumntype{L}{>{\raggedright\arraybackslash}X}
\newcommand{\bench}{S1CogBench}

\title{S1CogBench: What System One Decision Models Can and Cannot Do on Cognitive Experiments}
\author{\name Dengzhe Hou\textsuperscript{*}\qquad Kazunori D Yamada \\
\addr Tohoku University\qquad\textsuperscript{*}Corresponding author \\
\normalfont\small
\href{https://github.com/dengzhe-hou/S1CogBench}{\raisebox{-0.17em}{\includegraphics[height=1.05em]{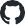}}\,GitHub}\qquad
\href{https://huggingface.co/datasets/dengzhe-hou/S1CogBench}{\raisebox{-0.17em}{\includegraphics[height=1.05em]{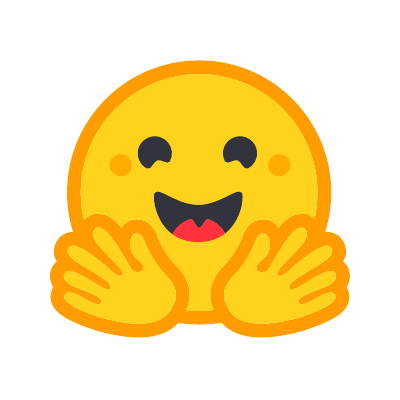}}\,Hugging Face}\qquad
\href{https://dengzhe-hou.github.io/S1CogBench/}{\raisebox{-0.17em}{\includegraphics[height=1.05em]{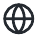}}\,Homepage}}
\newcommand{\halffigwidth}{0.62\linewidth}
\newcommand{\coveragesize}{\footnotesize}
\begin{document}
\raggedbottom
\maketitle
\begin{abstract}
Decision-making in humans and AI involves multiple cognitive abilities. System One decision models produce fast, structured choices. Their cognitive strengths and limitations, however, remain poorly understood. We introduce S1CogBench, a benchmark of 24 established cognitive tasks and 27 evaluation modes, evaluating seven open models and the API model Jev. It combines human-response prediction, autonomous execution, and controlled task manipulations to reveal strengths and failures that aggregate scores miss. We find distinct behavioral profiles across the evaluated models. In these models, strong performance on some learning tasks coexists with failures to gather information or adapt to feedback, while high accuracy does not guarantee human-like responses to changing task demands. Shared readout controls show that model rankings depend partly on how choices are extracted. Separate latency comparisons with generative language models show that long responses are much slower than direct decisions, whereas matched one-token generation adds little overhead. To our knowledge, S1CogBench is the first benchmark of System One decision models on established cognitive experiments. It provides a common empirical basis for identifying their capabilities and limitations before integrating them into automated decision systems.

\end{abstract}
\section{Introduction}
\label{sec:introduction}
Jev, Laya, and Intern-Decision exemplify a shift from free text generation to structured decisions \citep{typesafeJev,convaiLaya,internlm2026decision}. They return one typed decision over the legal options in a single call, so software can act on it directly for routing, guardrails, and automated gates, and a small model of this kind can be invoked many times inside a larger system. We call them \emph{System One decision models} \citep{typesafeJev}, also called typed decision models \citep{sharma2026typedbench}, as an operational label. We examine their cognitive strengths and limitations across established experiments.

\begin{figure*}[!t]
\vspace*{-10pt}
\centering
\setlength{\abovecaptionskip}{4pt}
\makebox[\textwidth][c]{\includegraphics[width=\textwidth,pagebox=cropbox]{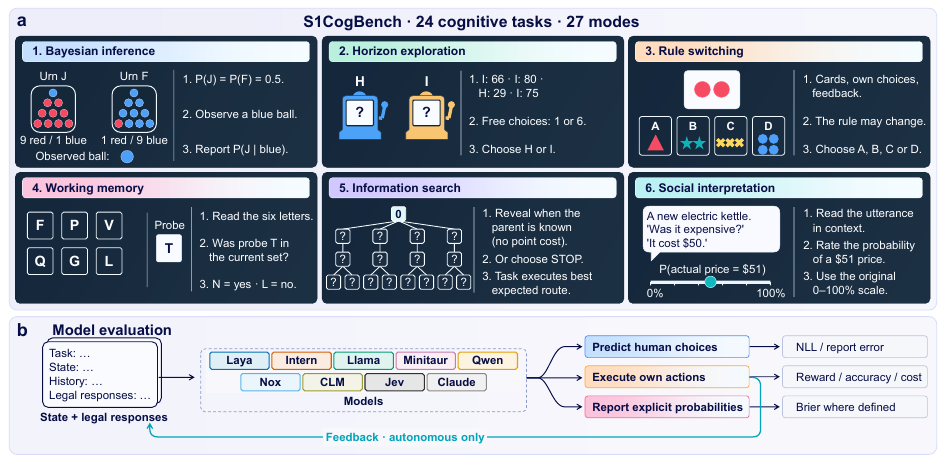}}
\caption{\textbf{S1CogBench connects cognitive tasks to model evaluation.}
\textbf{(a)} Six representative tasks from the 24-task, 27-mode suite.
\textbf{(b)} Adapters supply visible states, histories, and legal responses to the models, and each output has its own score. Model-specific coverage and API configurations are recorded in Appendix~\ref{app:selected-api-extension}.}
\label{fig:framework}
\vspace{-12pt}
\end{figure*}

A typed interface makes a response easy to execute, but it does not show that the response uses the relevant evidence, and no generated text is left to inspect \citep{sharma2026typedbench}. A model can return a well-formed distribution yet ignore feedback or stop searching too early. In closed-loop environments small System One models rarely succeed \citep{chen2026system1arena}, but such environments do not isolate which demand fails. Cognitive experiments can, because they manipulate information, uncertainty, memory load, incentives, or future opportunities and show how behavior changes.
Existing cognitive benchmarks evaluate language and multimodal models \citep{codaforno2024cogbench,binz2025centaur,ying2026coggym,nie2025evolve,hou2026cogbattery}, and CogGym's 50 models include no System One decision model. System1-Arena does evaluate System One decision models, but on static decision tasks and agent environments \citep{chen2026system1arena}. To our knowledge, \bench{} is the first benchmark to evaluate System One decision models on established cognitive experiments.

We ask what these models can and cannot do on these experiments. \textbf{RQ1} asks whether a model that acts well also predicts humans. \textbf{RQ2} asks what overall scores hide about how behavior changes across conditions. \textbf{RQ3} asks how latency varies across decision protocols.
Three findings characterize the cognitive behavior of the evaluated models.
\begin{itemize}[leftmargin=*,itemsep=1pt,topsep=2pt]
\item Learning curves differ substantially across models. In the reinforcement learning and working memory (RLWM) task, Intern-Decision improves across repeated presentations, whereas Laya remains almost flat (Section~\ref{sec:rq2}).
\item Autonomous performance does not guarantee accurate human-response prediction. In the Wisconsin Card Sorting Test (WCST), Nox has the highest autonomous accuracy among the open models but predicts human choices worse than uniform choice, whereas Minitaur predicts them best (Section~\ref{sec:rq1}).
\item Overall accuracy can hide differences in sensitivity to task demands. Nox achieves high RLWM accuracy but shows a much smaller memory-load effect than humans. In Horizon, each of the eight models responds less than humans to increased future opportunities, with participant-paired intervals below zero (Section~\ref{sec:rq2}).
\end{itemize}
Additional controls examine how readout rules affect model rankings and how response protocols affect latency (Sections~\ref{sec:rq1} and~\ref{sec:rq3}).

\bench{} is built for models that return a fast decision in a single call. It turns each of 24 tasks into a typed request over legal options, scores option probabilities and native choices, times each decision with generation as a control, and supports new model adapters. Figure~\ref{fig:framework} summarizes this connection.

\section{Related work}
\label{sec:related}
Cognitive benchmarks evaluate language and multimodal models. CogBench uses ten behavioral metrics from seven experiments \citep{codaforno2024cogbench}, CogGym evaluates 50 models on 258 single-turn experiments \citep{ying2026coggym}, and Psych-101 and Centaur evaluate held-out human-response prediction and feed model-generated choices back into Horizon and Two-step \citep{binz2025centaur}. CogBattery compares 54 language models with human data on 13 cognitive paradigms \citep{hou2026cogbattery}, and WMIP probes cumulative state tracking \citep{hou2026wmip}. \citet{namazova2025centaur} compare predictive and generative behavior, and \bench{} extends this distinction to explicit event probabilities, source-defined condition effects, and structured interfaces.

System1-Arena compares System One decision models through typed probability requests, closed-loop environments, latency, and cost \citep{chen2026system1arena}, and TypedBench examines calibration, wording and option sensitivity, length, and asymmetric decision costs \citep{sharma2026typedbench}. These works already provide unified interfaces, conditional diagnostics, and efficiency measurement. Table~\ref{tab:direct-benchmark-comparison} compares the closest resources with ours. Further related work appears in \hyperref[app:related]{Appendix~\ref*{app:related}}.

\begin{table*}[!t]
\providecommand{\coveragesize}{\small}
\centering\coveragesize
\begin{tabular}{@{}lccccc@{}}
\toprule
Resource & \begin{tabular}[b]{@{}c@{}}Cognitive\\experiments\end{tabular} & \begin{tabular}[b]{@{}c@{}}Predicts human\\responses\end{tabular} & \begin{tabular}[b]{@{}c@{}}Closed-loop\\actions\end{tabular} & \begin{tabular}[b]{@{}c@{}}System One\\models\end{tabular} & \begin{tabular}[b]{@{}c@{}}Decision time\\or cost\end{tabular} \\
\midrule
CogBench \citep{codaforno2024cogbench} & 7 & -- & $\checkmark$ & -- & -- \\
CogGym, Sept.~2026 v1 \citep{ying2026coggym} & 258 & $\checkmark$ & -- & -- & -- \\
Psych-101 / Centaur \citep{binz2025centaur} & 160 & $\checkmark$ & ($\checkmark$) & -- & -- \\
CogBattery \citep{hou2026cogbattery} & 13 & -- & -- & -- & -- \\
BanditBench / EVOLvE \citep{nie2025evolve} & -- & -- & $\checkmark$ & -- & -- \\
System1-Arena \citep{chen2026system1arena} & -- & -- & $\checkmark$ & $\checkmark$ & $\checkmark$ \\
TypedBench \citep{sharma2026typedbench} & -- & -- & -- & $\checkmark$ & $\checkmark$ \\
\midrule
\rowcolor{tablegray} \textbf{\bench{}} & \textbf{24} & $\checkmark$ & $\checkmark$ & $\checkmark$ & $\checkmark$ \\
\bottomrule
\end{tabular}
\caption{Coverage of the closest evaluation resources in each paper's main evaluation. Counting units differ across resources, and ($\checkmark$) marks selected tasks only. This compares coverage, not scores. The gray row highlights \bench{}.}
\label{tab:direct-benchmark-comparison}
\vspace{-6pt}\end{table*}

\section{The \bench{} benchmark}
\label{sec:benchmark}
\bench{} maps an original cognitive experiment to the information visible at each decision, its legal responses, a model output, a readout, and source-grounded scores.

\subsection{Task selection and provenance}
The 24 tasks fall into ten organizational domains (Table~\ref{tab:domains}). WCST, Horizon, and Mouselab each have a human-prediction and an autonomous mode, giving 27 modes. The task-source catalogue in the project repository documents the source of each task. Each specification identifies the original study, adopted public dataset or environment version, manipulation, stimuli, legal responses, feedback, reset and stopping rules, and scoring unit. The 24 tasks come from published experiments and benchmarks, such as Horizon \citep{wilson2014exploration}, WCST \citep{grant1948cardsorting}, RLWM \citep{collins2012rlwm}, Mouselab \citep{callaway2022planning}, and the two-step task \citep{daw2011model}. Official splits are used where provided.

\begingroup
\setlength{\intextsep}{6pt}
\begin{table*}[!htbp]
\centering\footnotesize
\begin{tabularx}{\linewidth}{@{}lLL@{}}
\toprule
Domain & Tasks & What the experiments vary \\
\midrule
\textcolor[rgb]{0.078431,0.341176,0.639216}{Evidence} & \textbf{1}~Bayesian urn, \textbf{2}~multiple-cue judgment, \textbf{3}~multi-attribute choice, \textbf{4}~probability coherence & Prior probability, likelihood, cue validity, coherence \\
\addlinespace[2pt]
\textcolor[rgb]{0.552941,0.454902,0.066667}{Categories} & \textbf{5}~SHJ, \textbf{6}~Medin, \textbf{7}~weather prediction & Category structure, feedback, transfer and typicality \\
\addlinespace[2pt]
\textcolor[rgb]{0.639216,0.078431,0.337255}{History} & \textbf{8}~RLWM, \textbf{9}~recent probes & Set size, repetition interval, recent interference \\
\addlinespace[2pt]
\textcolor[rgb]{0.639216,0.341176,0.078431}{Adaptation} & \textbf{10}~WCST, \textbf{11}~restless bandit, \textbf{12}~instrumental learning & Rule changes, changing rewards, outcome feedback \\
\addlinespace[2pt]
\textcolor[rgb]{0.109804,0.513725,0.376471}{Exploration} & \textbf{13}~Horizon, \textbf{14}~stationary bandits & Future opportunities, uncertainty \\
\addlinespace[2pt]
\textcolor[rgb]{0.172549,0.078431,0.639216}{Information} & \textbf{15}~Hybrid sampling, \textbf{16}~Mouselab & Reveal actions, sampling cost, stopping conditions \\
\addlinespace[2pt]
\textcolor[rgb]{0.592157,0.376471,0.125490}{Preference} & \textbf{17}~BART, \textbf{18}~temporal discounting & Risk, reward magnitude, explicit delays \\
\addlinespace[2pt]
\textcolor[rgb]{0.125490,0.258824,0.592157}{Planning} & \textbf{19}~Two-step & Transitions, changes in outcome value \\
\addlinespace[2pt]
\textcolor[rgb]{0.090196,0.513725,0.529412}{Social} & \textbf{20}~False-belief QA, \textbf{21}~politeness, \textbf{22}~pragmatic interpretation, \textbf{23}~nonliteral number & Belief access, communicative goals, literal meaning \\
\addlinespace[2pt]
\textcolor[rgb]{0.380392,0.380392,0.380392}{Metacognition} & \textbf{24}~Self-assessment & Difficulty, correctness \\
\bottomrule
\end{tabularx}
\caption{The 24 tasks in ten organizational domains. Domains organize the tasks and are not validated latent factors. Task sources are listed in the task-source catalogue in the project repository.}
\label{tab:domains}
\vspace{-6pt}\end{table*}
\endgroup

\subsection{Typed decision interface}
For each decision, an adapter gives the model what the original task shows at that point, the history $h_t$ as a visible state $x_t$, and the ordered legal responses $\mathcal A_t$. For human prediction the history is the participant's, and for autonomous execution it is the model's own. Histories can be complete, summarized, windowed, or absent, and these are different information conditions. Responses can be categorical actions, ordinal ratings, numerical reports, or event probabilities.

We evaluate the released models without further fitting to the evaluation participants. In sequential human-response prediction, each model uses that participant's preceding history in the form supplied by the task adapter. Conditioning on this history does not update model weights. The fitting data and evaluation splits of cognitive references are documented separately in Appendix~\ref{app:protocol}, including cases without held-out participants.

The model returns a distribution $p_\theta(a\mid x_t,\mathcal A_t)$ over those responses. A readout is the rule that turns this output into a probability or an executed action, such as a native choice, a sampled action, or generative decoding. We record the distribution, the native choice, the executed action, the feedback, and the timing, so a model's distribution stays separate from randomness added at execution. An episode ends as in the original task, for example after a fixed number of trials, an explicit stop, or a terminal state.

We document the adaptations to task presentation, supplied histories, and legal response options. Hybrid sampling supplies the authors' posterior summary. In autonomous WCST, rules switch according to the source timeline, independently of model actions. In autonomous Mouselab, observing nodes has no cost. The environment selects the final route with the highest expected value. Stationary bandits provide choice counts and mean rewards from the original implementation instead of the full trial history.

\subsection{Scoring targets}
For observed participant $i$ and scored trial $t$, human-response NLL scores the observed choice given its preceding history,
\begin{equation}
 \operatorname{NLL}=-\sum_{i,t}w_{it}\log p_\theta(a_{it}\mid h_{it}),\qquad \sum_{i,t}w_{it}=1.
 \label{eq:nll}
\end{equation}
The weights follow the source scorer. For $N$ equally weighted responses, $w_{it}=1/N$. For an equal mean over $M$ participants, each with $n_i$ scored responses, $w_{it}=1/(M n_i)$. Fold-level aggregation follows the source protocol (Appendix~\ref{app:scoring-reproduction}).
NLL over the legal responses and NLL over the full vocabulary are different scores. A returned zero gives infinite NLL unless the primary metric of the source scorer defines an epsilon, as in WCST and Horizon.

Autonomous scores use executed trajectories, with total reward
\begin{equation}
 G=\sum_t(r_t-c_t),
 \label{eq:return}
\end{equation}
where the reward, each cost, and the horizon follow the source environment. Single-event Brier $(q-Y)^2$ needs an explicit probability and an observed event. Answer-category Brier sums over categories, so the two are not on the same scale.

We report nine shared measures, accuracy, reward, pseudo-regret, MAE, human NLL, Brier, latency P50 and P95, and invalid-response rate, each where the task defines it. Per-mode definitions and tables, complete curves, per-target API records, and reference details are in the project repository.{} Appendix~\ref{app:scoring-reproduction} specifies the primary probability scores and timing boundaries.\ Original cognitive effects are reported alongside these scores. Uncertainty uses participants or independent environments, not trajectory steps, and paired 95\% intervals state their resampling unit.

Uniform assigns equal probability to each legal response. Reference is a task-specific baseline based on human responses, fitted models, policies, analytic solutions, or answer keys (Appendix~\ref{app:protocol}).

\section{Experiments}
\label{sec:results}
\begin{table*}[!t]
\centering\footnotesize
\setlength{\tabcolsep}{0pt}
\begin{tabular*}{\linewidth}{@{\extracolsep{\fill}}l r rrrrrrr@{\hspace{12pt}}rr@{\hspace{12pt}}r@{}}
\toprule
 & & \multicolumn{7}{c}{Open models} & \multicolumn{2}{c}{API models} & \\
\cmidrule(lr){3-9}\cmidrule(lr){10-11}
Task & Uniform & Laya & Intern & Llama & Minitaur & Qwen & Nox & CLM & Jev & Haiku & Reference \\
\midrule
\multicolumn{12}{@{}>{\columncolor{tablegray}}l}{\textit{Human-response NLL $\downarrow$}} \\
Multiple-cue judgment & 2.197 & \textcolor{gray}{2.342} & \underline{1.570} & \textcolor{gray}{2.687} & 1.984 & \textcolor{gray}{3.012} & \textbf{1.491} & \textcolor{gray}{2.675} & $+\infty$ & -- & 2.029 \\
Multi-attribute choice & 0.693 & \textcolor{gray}{0.730} & \underline{0.546} & \textcolor{gray}{0.786} & \textcolor{gray}{0.722} & 0.613 & \textbf{0.366} & \textcolor{gray}{1.039} & 0.329 & 0.206 & 0.192 \\
WCST & 1.386 & \textcolor{gray}{1.462} & \underline{1.378} & \textcolor{gray}{1.431} & \textbf{1.344} & \textcolor{gray}{1.968} & \textcolor{gray}{1.552} & \textcolor{gray}{2.368} & 1.543 & -- & 0.840 \\
Horizon & 0.693 & \textcolor{gray}{0.702} & 0.518 & 0.503 & \underline{0.409} & 0.661 & \textbf{0.408} & \textcolor{gray}{1.161} & 0.414 & 0.390 & 0.400 \\
Hybrid sampling & 1.386 & \textcolor{gray}{1.390} & \underline{1.116} & \textcolor{gray}{1.515} & \textbf{1.039} & \textcolor{gray}{2.648} & \textcolor{gray}{2.044} & \textcolor{gray}{3.012} & $+\infty$ & -- & 0.264 \\
Mouselab & 1.634 & \textcolor{gray}{\underline{1.717}} & \textcolor{gray}{2.109} & \textcolor{gray}{2.743} & \textcolor{gray}{2.125} & \textcolor{gray}{2.464} & \textbf{1.619} & \textcolor{gray}{7.817} & 1.458 & -- & 1.367 \\
Politeness & 1.014 & \textcolor{gray}{1.192} & \underline{0.799} & \textcolor{gray}{1.584} & 1.008 & 1.000 & \textbf{0.749} & \textcolor{gray}{1.195} & $+\infty$ & $+\infty$ & 0.587 \\
Pragmatic interpretation & 1.282 & \textcolor{gray}{1.592} & \textbf{0.814} & \textcolor{gray}{1.851} & 1.049 & 1.240 & \underline{0.895} & 1.081 & $+\infty$ & $+\infty$ & 0.416 \\
Self-assessment & 2.039 & \textcolor{gray}{$+\infty$} & 2.029 & \textcolor{gray}{2.523} & \textbf{1.885} & \textcolor{gray}{2.262} & \underline{1.893} & \textcolor{gray}{3.094} & $+\infty$ & -- & 1.619 \\
\midrule
\multicolumn{12}{@{}>{\columncolor{tablegray}}l}{\textit{Accuracy under given histories (\%) $\uparrow$}} \\
SHJ categories & 50.0 & \textcolor{gray}{50.0} & 65.0 & 52.9 & 54.5 & \underline{71.3} & \textbf{94.5} & 50.4 & 87.4 & 91.2 & 93.8 \\
Medin categories & 50.0 & \textcolor{gray}{49.6} & \underline{72.3} & 53.5 & 55.9 & 53.8 & \textbf{92.8} & 50.4 & 86.8 & 97.4 & 96.6 \\
Weather prediction & 50.0 & 54.5 & \underline{73.0} & \textcolor{gray}{47.5} & 53.8 & 50.5 & \textbf{81.2} & 52.5 & 80.8 & 80.2 & 78.2 \\
RLWM & 33.3 & 34.9 & \underline{77.8} & 41.7 & 49.0 & 59.3 & \textbf{86.3} & 36.7 & 89.5 & 89.4 & 89.3 \\
Recent probes & 50.0 & \textcolor{gray}{49.7} & \underline{82.1} & 55.6 & \textcolor{gray}{49.5} & 53.5 & \textbf{99.3} & 55.9 & 99.0 & 100.0 & 100.0 \\
False-belief QA & 50.0 & \textcolor{gray}{39.5} & \underline{82.5} & 81.8 & 65.0 & \textbf{82.8} & 81.9 & 67.4 & 71.5 & 82.6 & 100.0 \\
\midrule
\multicolumn{12}{@{}>{\columncolor{tablegray}}l}{\textit{Autonomous choices, correct or optimal (\%) $\uparrow$}} \\
WCST & 25.0 & 25.5 & 27.9 & \underline{31.4} & 28.2 & 25.2 & \textbf{32.1} & 25.7 & 28.5 & -- & 48.9 \\
Restless bandit & 50.0 & 55.4 & \underline{65.6} & 57.1 & \textcolor{gray}{49.0} & 50.5 & \textbf{69.0} & 52.6 & 63.9 & -- & 68.8 \\
Instrumental learning & 75.0 & \textcolor{gray}{73.9} & 81.4 & 79.2 & \underline{86.7} & 80.0 & \textbf{89.8} & \textcolor{gray}{69.5} & 86.7 & -- & 94.9 \\
Horizon & 50.0 & 54.8 & \underline{78.7} & 66.4 & 71.7 & 63.3 & \textbf{86.0} & \textcolor{gray}{50.0} & 82.9 & -- & 74.6 \\
Stationary bandits & 12.5 & 17.6 & \textbf{37.4} & 13.5 & \textcolor{gray}{9.8} & 28.9 & \underline{35.4} & 15.4 & 47.0 & -- & 68.1 \\
\midrule
\multicolumn{12}{@{}>{\columncolor{tablegray}}l}{\textit{Autonomous reward $\uparrow$}} \\
Mouselab & 7.01 & \textcolor{gray}{0.00} & \textcolor{gray}{0.00} & \textcolor{gray}{\underline{3.77}} & \textcolor{gray}{0.00} & \textbf{17.40} & \textcolor{gray}{\underline{3.77}} & \textcolor{gray}{0.00} & 8.21 & -- & 11.77 \\
BART & 0.84 & \textcolor{gray}{0.00} & \textcolor{gray}{0.74} & \textcolor{gray}{0.28} & \textcolor{gray}{0.00} & \textcolor{gray}{0.00} & \textbf{4.72} & \underline{1.03} & 1.76 & 6.27 & 10.49 \\
Two-step & 0.50 & 0.50 & \underline{0.52} & 0.50 & 0.51 & 0.52 & \textbf{0.53} & \textcolor{gray}{0.49} & 0.55 & 0.55 & 0.52 \\
\midrule
\multicolumn{12}{@{}>{\columncolor{tablegray}}l}{\textit{Report error (MAE) $\downarrow$}} \\
Bayesian urn & 0.349 & 0.291 & \underline{0.235} & 0.240 & 0.294 & 0.246 & \textbf{0.175} & 0.286 & 0.068 & 0.001 & 0.000 \\
Probability coherence & 32.13 & \textcolor{gray}{32.98} & \underline{18.95} & 28.57 & 19.04 & 21.87 & \textbf{18.03} & \textcolor{gray}{42.13} & 17.66 & -- & 16.54 \\
Nonliteral number & 34.77 & 30.86 & \textbf{16.87} & 25.10 & 28.53 & 29.75 & \underline{17.73} & 18.97 & 20.37 & 22.52 & 4.98 \\
\midrule
\multicolumn{12}{@{}>{\columncolor{tablegray}}l}{\textit{Preference, later reward chosen (\%)}} \\
Temporal discounting & 50.0 & 0.0 & 16.7 & 0.0 & 80.0 & 90.0 & 16.7 & 0.0 & 50.0 & 77.8 & 41.5 \\
\bottomrule
\end{tabular*}
\caption{Primary scores across 27 modes. Arrows indicate the better direction. Bold and underlining mark the best and second-best unrounded scores among seven open models, including ties, without implying significance. Gray text indicates no improvement over Uniform, a dash no complete score, and $+\infty$ zero probability on a human choice without epsilon. Appendices~\ref{app:protocol}, \ref{app:table2-controls}, and~\ref{app:scoring-reproduction} give protocol details, statistical support, and scoring definitions.}
\label{tab:task-summary}
\end{table*}

\subsection{Setup}
\label{sec:protocol}
We evaluate seven open models as released. Laya, a non-autoregressive model, uses a bidirectional ModernBERT encoder with a learned decision head~\citep{convaiLaya}. Intern-Decision-4B is fine-tuned from Qwen3.5-4B~\citep{qwen35_4b} for structured decisions while retaining its causal backbone and native language-model head~\citep{internlm2026decision}. Decision-2.0-Nox-4B~\citep{vllmsr2026nox} is fine-tuned from Qwen3.5-4B-Base. CLM-v0.1-8B~\citep{contrastivelm2026clm} scores options by the similarity between state and option embeddings produced by two trained heads on a frozen Qwen3-8B encoder~\citep{qwen3_8b}. These four are System One decision models. Qwen3.5-4B and Llama-3.1-8B-Instruct~\citep{grattafiori2024llama3} serve as general-purpose controls, while Minitaur-8B~\citep{binzMinitaur} is fine-tuned on human behavior from Psych-101.

Intern-Decision and Qwen3.5-4B form a backbone comparison. Nox does not form such a pair with the post-trained Qwen3.5-4B evaluated here, and we do not separately evaluate CLM's Qwen3-8B encoder. Training data and compute are not matched across models. Minitaur has seen other participants from the experiments underlying Horizon and the eight modes drawn from the Psych-101 test split. Its human-prediction scores on these tasks therefore represent in-distribution performance.

All 168 combinations of task and open model use the full sample. Only multiple-cue judgment uses a compact history, for Laya, Intern-Decision, and Qwen. Jev runs all 27 modes through its API, while Claude5.5, a general model with adaptive thinking, runs selected tasks because of cost (Appendix~\ref{app:selected-api-extension}). On Horizon and WCST, we additionally compare legal argmax and categorical sampling across all seven open models.

\subsection{Overall performance}
\label{sec:main-results}
Table~\ref{tab:task-summary} reports one primary target for each of the 27 modes. We compare performance on 26 targets, as temporal discounting measures a preference for later rewards rather than better performance. Under their original readouts, Laya does not outperform Uniform on 17 targets, CLM on 13, Llama on 11, Minitaur on 7, Qwen on 6, and Intern-Decision and Nox on 3 each. Nox leads the seven open models on 18 targets.

With 10,000 bootstrap resamples over participants, episodes, or items as appropriate, 95\% intervals support performance above Uniform on 7 targets for Laya, 11 for CLM, 21 for Intern-Decision, and 23 for Nox (Table~\ref{tab:uniform-intervals}). Excluding the three targets with at most six sampling units and additionally requiring the three report-error targets to beat a constant midpoint report (0.5 or 50) gives counts of 4, 9, 18, and 20 of 23, respectively.

Paired by sampling unit, Intern-Decision reliably outperforms its Qwen3.5 backbone on 19 targets and performs worse on 2. Excluding the three few-unit targets gives 17 better and 2 worse results out of 23.

Two cells of Table~\ref{tab:task-summary} reflect interface failures. Laya gives zero probability to 192 of 108,248 human responses in self-assessment, because its SDK rounds probabilities to four decimals, so its NLL there is infinite. Its true NLL is at least about 2.48, still worse than uniform. Minitaur reports 50\% on 319 of 330 nonliteral-number items, a near-constant answer.

\subsection{Does acting well imply predicting humans? (RQ1)}
\label{sec:rq1}
\begin{figure*}[t]
\centering
\includegraphics[width=\textwidth]{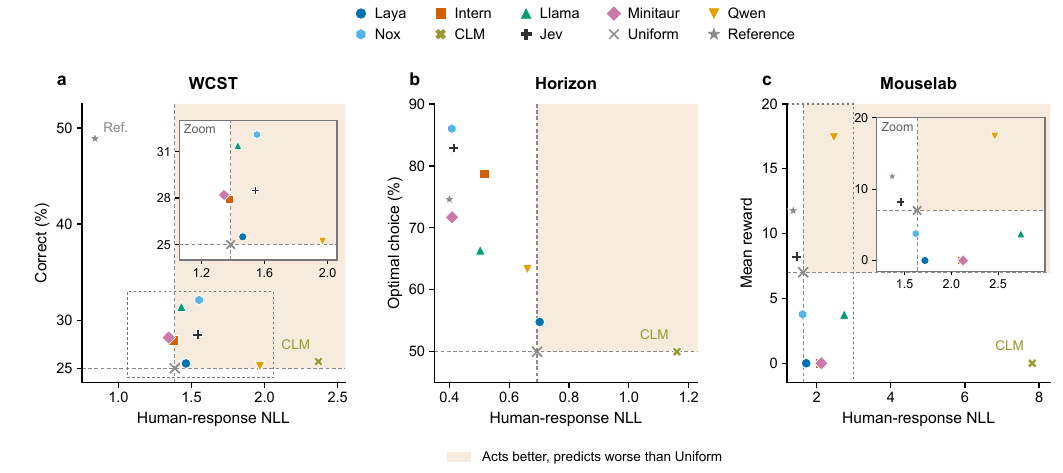}
\caption{Human prediction against autonomous performance under the original readouts, from Table~\ref{tab:task-summary}. Dashed lines mark Uniform, and shading indicates better autonomous performance but worse human prediction. Main axes show all complete score pairs. Insets magnify the dashed rectangles in panels a and c. The axes summarize separate evaluation modes rather than paired decisions.}
\label{fig:rq1-targets}
\end{figure*}

Autonomous performance and human-response prediction can rank models differently. Their relationship depends on the task and the rule used to select actions. WCST separates the two evaluations under the same task rules (Figure~\ref{fig:rq1-targets}a). Under the original readouts, Nox has the highest autonomous accuracy among the open models (32.1\%) but predicts human choices worse than Uniform (NLL 1.552 versus 1.386), whereas Minitaur predicts them best (NLL 1.344). The action rule also matters. When all models select their highest-probability legal option, Minitaur also has the highest accuracy (34.3\%). Sampling leaves CLM near Uniform in both Horizon and WCST (Tables~\ref{tab:p3-final-horizon} and~\ref{tab:wcst-readouts}).

Horizon instead shows agreement between the two evaluations (Figure~\ref{fig:rq1-targets}b). Nox leads the seven open models in both human prediction (NLL 0.408) and autonomous reward (56.21), and retains the highest reward under shared argmax and sampling (Table~\ref{tab:p3-final-horizon}). The RW reference, fitted to 25 training participants, predicts human choices slightly better (NLL 0.400) but earns less reward (54.56).

Mouselab also separates the two, but its autonomous mode removes the observation cost that participants paid, so revealing more nodes pays off (Figure~\ref{fig:rq1-targets}c). Qwen reveals 15.5 nodes per episode and earns the highest autonomous reward among the open models (17.40), close to the 17.55 obtained by revealing every node and above the 11.77 of MetaGreedy, a one-step value-of-information policy replayed on the same trees (Appendix~\ref{app:new-traditional-references}), yet it predicts human choices worse than uniform choice (NLL 2.464 versus 1.634). Jev reveals 3.3 nodes and earns 8.21. Nox and Llama reveal one node and stop, so both earn exactly 3.77, below Uniform's 7.01, and Nox's human NLL is only slightly better than Uniform (1.619). Intern, Laya, Minitaur, and CLM stop immediately in all 23,750 episodes and receive zero reward, although Laya ranks second in human prediction (NLL 1.717). STOP is always the first listed option, so this stopping may partly reflect a position bias.

The paired tasks compare acting with predicting humans. We now turn to the quality of the probabilities models report. In Bayesian urn inference, Nox leads all seven open models on posterior MAE (0.1748) and event Brier error (0.1990) over 400 queries. Intern ranks second on MAE (0.2345), while Qwen ranks second on Brier error (0.2299).

\begin{table}[!htbp]
\centering\footnotesize
\setlength{\tabcolsep}{1.6pt}
\begin{tabular}{@{}lrrrrrrr@{}}
\toprule
 & Human & \multicolumn{3}{c}{Mean reward $\uparrow$} & \multicolumn{3}{c}{Optimal choice (\%) $\uparrow$} \\
\cmidrule(lr){3-5}\cmidrule(lr){6-8}
Model & NLL $\downarrow$ & orig. & argmax & sample & orig. & argmax & sample \\
\midrule
\rowcolor{tablegray} Uniform & 0.693 & 50.0 & 50.0 & 50.0 & 50.0 & 50.0 & 50.0 \\
Laya & 0.702 & 51.0 & 51.8 & 51.0 & 54.8 & 57.9 & 54.8 \\
Intern & 0.518 & \underline{54.9} & 54.9 & 52.2 & \underline{78.7} & 78.7 & 62.8 \\
Llama & 0.503 & 52.9 & 53.6 & 52.7 & 66.4 & 70.9 & 65.2 \\
Minitaur & \underline{0.409} & 53.9 & \underline{56.2} & \underline{53.7} & 71.7 & \underline{84.0} & \underline{70.9} \\
Qwen & 0.661 & 52.2 & 52.2 & 51.8 & 63.3 & 63.3 & 60.8 \\
Nox & \textbf{0.408} & \textbf{56.2} & \textbf{56.2} & \textbf{55.4} & \textbf{86.0} & \textbf{86.0} & \textbf{80.7} \\
CLM & 1.161 & 49.8 & 49.8 & 50.0 & 50.0 & 50.0 & 50.7 \\
\rowcolor{tablegray} RW fit & 0.400 & 54.6 & 56.0 & 54.6 & 74.6 & 84.9 & 74.6 \\
\bottomrule
\end{tabular}
\caption{Horizon human prediction (six held-out participants, 6,720 free choices) and autonomous play (31 timelines, 34,720 free choices per model and readout). Arrows indicate the better direction. Orig. is each model's own readout (Appendix~\ref{app:protocol}), and argmax and sample apply legal argmax and legal sampling to all models. RW fit is a Rescorla-Wagner model fitted to 25 training participants. Bold and underlining mark each column's best and second-best open models before rounding. Gray rows denote the Uniform and RW fit references.}
\label{tab:p3-final-horizon}
\end{table}

Jev performs strongly in RLWM, reaching 89.5\% accuracy under human-block histories, close to the 89.3\% of the same-participant fitted reference under argmax. In autonomous Horizon, its optimal-choice rate is 82.9\%, between the human-fitted reference's 74.6\% under sampling and 84.9\% under greedy replay. For human prediction, Jev's two-decimal probabilities can assign zero to a person's actual choice. Its NLL is therefore infinite on five of nine prediction targets. Those five have no autonomous mode, so they do not supply a paired comparison. On tasks with both modes, Jev beats Uniform in Horizon and Mouselab (human NLL 0.414 and 1.458). In WCST, its NLL (1.543) is worse than Uniform and its accuracy (28.5\%) is close to it.

\FloatBarrier
\subsection{What do aggregate scores hide? (RQ2)}
\label{sec:rq2}
\begin{figure*}[!hbp]
\centering
\includegraphics[width=\textwidth]{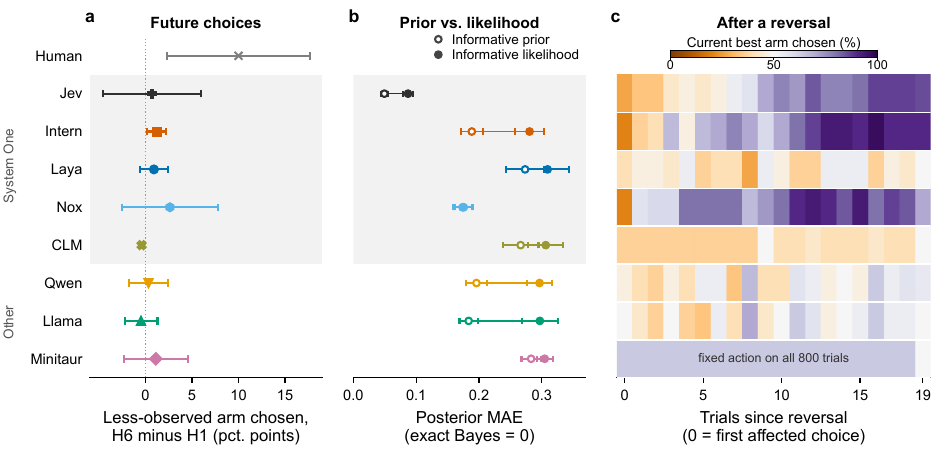}
\caption{Cognitive profiles. (a) Change in choosing the less-observed arm from horizon 1 to 6 at the first free choice after unequal information, observed for humans and predicted by models from the same histories. (b) Posterior MAE with a more informative prior or likelihood. (c) Choices aligned to reversals over 10 runs, with two reversals per run at trial 19 and three elsewhere. Minitaur chose arm F on all 800 trials, so its flat row is not adaptation. Error bars are 95\% Student-t intervals over six participants (a) and 200 paired seeds (b), uncorrected for multiple comparisons. Panels b and c have no human data.}
\label{fig:exploratory-cognitive-profiles}
\end{figure*}

Average scores can hide differences in exploration, rule switching, memory load effects, and the use of evidence and feedback. Horizon shows a response to future opportunities that the pooled NLL misses. When more choices remain (horizon 6 against 1), humans choose the less-observed arm 10.00 percentage points more often at their first unequal-information free choice. Under the same histories, models predict changes of only $-0.49$ to $2.62$ points (Figure~\ref{fig:exploratory-cognitive-profiles}a). Paired by participant, the model-minus-human difference has a 95\% interval below zero for all eight models.

In WCST, fewer perseveration errors do not necessarily indicate better rule switching. Laya makes fewer such errors than Llama (15.24\% versus 28.82\%), but its accuracy is also lower (25.47\% versus 31.39\%). Uniform choice has a similarly low perseveration rate (17.13\%), showing why this measure must be read alongside accuracy (Table~\ref{tab:wcst}). Nox shows the same pattern under sampling, with fewer perseveration errors and lower accuracy (Table~\ref{tab:wcst-readouts}).
\begin{table}[!t]
\begin{minipage}[t]{0.49\textwidth}
\vspace{0pt}
\centering\footnotesize
\setlength{\tabcolsep}{1.3pt}
\begin{tabular}{@{}lrrrrr@{}}
\toprule
 & \multicolumn{2}{c}{Human history} & \multicolumn{3}{c}{Autonomous} \\
\cmidrule(lr){2-3}\cmidrule(lr){4-6}
Model & Accuracy $\uparrow$ & NLL $\downarrow$ & Accuracy $\uparrow$ & Persev. $\downarrow$ & Set-loss $\downarrow$ \\
\midrule
\rowcolor{tablegray} Human & 50.00 & -- & 50.00 & 7.32 & 1.88 \\
\rowcolor{tablegray} Uniform & 25.00 & 1.3863 & 25.00 & 17.13 & 17.13 \\
Laya & 25.53 & 1.4616 & 25.47 & 15.24 & 18.16 \\
Intern & 28.90 & \underline{1.3781} & 27.90 & 21.25 & 15.84 \\
Llama & \textbf{32.56} & 1.4307 & \underline{31.39} & 28.82 & 13.78 \\
Minitaur & 30.06 & \textbf{1.3435} & 28.16 & 21.85 & 14.61 \\
Qwen & 25.07 & 1.9677 & 25.20 & 15.88 & 17.88 \\
Nox & \underline{30.10} & 1.5518 & \textbf{32.14} & 32.53 & 11.88 \\
CLM & 25.20 & 2.3678 & 25.71 & 14.96 & 18.08 \\
\rowcolor{tablegray} SL refit & 60.40 & 0.8402 & 48.92 & 9.47 & 4.63 \\
\bottomrule
\end{tabular}
\caption{WCST means across 75 held-out participants (12,754 trials per mode). Percentages use all trials. Persev. errors repeat the previous wrong rule. Set-loss errors leave the previous correct rule. SL (sequential learning) was fitted to 300 other participants. Bold and underlining mark the best and second-best open models in each accuracy or NLL column. Gray rows denote the Human, Uniform, and SL refit references. Appendix~\ref{app:scoring-reproduction} gives scoring details.}
\label{tab:wcst}

\end{minipage}\hfill
\begin{minipage}[t]{0.49\textwidth}
\vspace{0pt}
\centering\footnotesize
\setlength{\tabcolsep}{2pt}
\begin{tabular}{@{}lrrrr@{}}
\toprule
 & & E2E P50 & Ratio to & Largest \\
Config. & Requests & (ms) $\downarrow$ & Llama $\downarrow$ & P95 (ms) $\downarrow$ \\
\midrule
Laya FP32 & 5,576 & 25.50--274.44 & 0.67--0.95 & 742.54 \\
Laya BF16 & 5,576 & 26.55--82.18 & 0.23--0.71 & 200.76 \\
Laya fast & 1,200 & 6.45--59.83 & 0.12--0.17 & 50,395 \\
Intern FLA & 5,576 & 52.06--219.40 & 0.75--1.41 & 437.76 \\
Intern Torch & 1,200 & 174.58--655.74 & 2.19--4.70 & 1,223 \\
Qwen FLA & 5,576 & 51.66--211.88 & 0.71--1.40 & 427.38 \\
Qwen Torch & 1,200 & 162.25--595.82 & 2.09--4.40 & 1,189 \\
Nox FLA & 1,200 & 52.97--232.12 & 0.80--1.43 & 510.72 \\
Nox Torch & 5,576 & 179.87--788.22 & 2.64--4.92 & 1,592 \\
CLM & 5,576 & 25.77--109.67 & 0.12--0.70 & 152.02 \\
\rowcolor{tablegray} Llama & 5,576 & 36.97--274.02 & 1.00 & 611.22 \\
Minitaur & 5,576 & 35.65--265.50 & 0.93--0.97 & 596.42 \\
\bottomrule
\end{tabular}
\caption{End-to-end (E2E) latency across six modes on one A100, batch size one. Ranges span modes. Ratios to Llama are medians over matched requests. The gray row marks Llama as the latency reference. FLA uses fused kernels. Torch runs without them. Laya fast's high P95 may reflect compilation. Appendix~\ref{app:scoring-reproduction} gives timing details.}
\label{tab:p1-latency-summary}

\end{minipage}
\end{table}

The text format alone does not explain these low accuracies. Under the original readouts, the seven open models score 25.20\% to 32.14\%, near Uniform (25\%) and below humans (50.00\%) and the sequential-learning reference (48.92\%). On 74 common decisions from two participants, Claude models with adaptive thinking score 63.51\% to 66.22\%, compared with 18.92\% to 43.24\% for open models under shared legal readouts (common-source comparison in the project repository). The readouts are not matched. This comparison shows that the text task still permits learning from feedback.

In RLWM, average accuracy does not predict sensitivity to memory load (Appendix Table~\ref{tab:rlwm-models}). Nox has high accuracy (86.30\%) but loses only 4.84 percentage points from set size 2 to 6, less than half the human effect (12.74 points). Intern and Minitaur have similar load effects, close to the human effect, despite their different average accuracies. Qwen shows a larger load effect, whereas CLM shows almost none. An observer that uses the same human histories without forgetting has no load effect ($+0.94$ points), so the human effect reflects a memory cost rather than the histories themselves.

\begin{figure}[!b]
\centering
\providecommand{\halffigwidth}{\columnwidth}
\includegraphics[width=0.7948717948717948\linewidth]{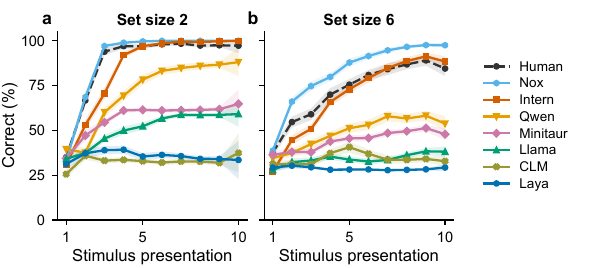}
\caption{RLWM correct responses by stimulus presentation under complete human-block histories, with pointwise 95\% Student-t intervals across participants (34 at set size 2 and 69 at set size 6 by presentation 10).}
\label{fig:rlwm-curves}
\end{figure}

The learning curves reveal a different contrast (Figure~\ref{fig:rlwm-curves}). From presentation 1 to 10 at set size 6, Intern gains 61.49 points and Nox gains 58.65, compared with 46.35 for humans. Intern starts below humans and ends above them. Laya and CLM remain nearly flat. The same observer gains 66.67 points, so a steep curve shows that a model uses the supplied history, not by itself that it learns as people do. Strong overall performance, sensitivity to memory load, and learning from repeated feedback therefore describe different aspects of behavior.

In Bayesian urn inference, six of the seven open models have higher MAE when the likelihood is more informative (0.28--0.31) than when the prior is more informative (0.18--0.28). Nox has similar errors in both conditions (0.174 and 0.176, Figure~\ref{fig:exploratory-cognitive-profiles}b).

In the Restless bandit, models must change their choices when the better arm reverses. Minitaur selects arm F on all 800 trials. Although its reported probability of F varies from 0.5622 to 0.9325, its action never changes. Changes in its heatmap row therefore reflect the changing environment rather than behavioral adaptation (Figure~\ref{fig:exploratory-cognitive-profiles}c).

Jev, Intern, and Nox choose the new best arm more often later after a reversal. Comparing the first five trials with trials 15--19, the rate rises from 36.7\% to 85.0\% for Jev, from 42.0\% to 91.3\% for Intern, and from 55.3\% to 82.7\% for Nox. The other models remain less successful in the later window, with rates of 46.0\% for Laya, 54.7\% for Llama, 55.0\% for Qwen, and 42.7\% for CLM.

\subsection{How does latency vary across decision protocols? (RQ3)}
\label{sec:rq3}
Under the same readout, the size-matched Intern-Decision and Qwen3.5 pair shows no speed advantage from fine-tuning for structured decisions. We time 49,408 requests across twelve configurations and 72 task-mode cells on one A100 at batch size one (Table~\ref{tab:p1-latency-summary}).

Latency also varies with input handling and runtime configuration. Longer inputs generally take more time, except for CLM, whose server truncates long inputs and caches shared prefixes. Kernel and precision choices also matter, as the Laya and Nox configurations show. Laya is much smaller than the other models (about 0.42 billion parameters), and its fast path has low medians but a recent-probes P95 of 50.4 seconds.

Extended free-text responses are much slower than direct decisions in our measurements. In Horizon, median generation takes 2.958 seconds for Llama and 16.330 seconds for Qwen under a 4,096-token cap, versus 36.97 and 51.66 milliseconds for single forward passes, about 80 and 316 times as long. The extra time does not guarantee better decisions. In WCST, Llama's free-text protocol under a 1,024-token cap scores 25.9\% over all 12,754 actions, including 2,217 invalid responses, against 31.4\% under its original readout. Qwen reaches the 4,096-token cap on 75 of 130 attempts. The protocols differ in prompts, lengths, histories and cache use, so these ratios do not isolate an interface effect.

Under matched inputs, greedy one-token generation takes 1.0024--1.0202 times as long as a legal-choice forward pass for Llama and Qwen. This control measures latency, and one generated token does not necessarily constitute a valid answer (Appendix~\ref{app:scoring-reproduction}).

\section{Discussion}
\label{sec:discussion}
The learning results raise a question about how different model designs use feedback. Laya reads inputs through a bidirectional encoder and a decision head~\citep{convaiLaya}, whereas Intern-Decision retains a causal language-model backbone~\citep{internlm2026decision}. Their contrasting RLWM curves suggest that returning a structured choice and extracting reward contingencies from a supplied history are distinct demands. Intern and Nox also differ in their sensitivity to memory load despite belonging to the same backbone family. These patterns motivate examining history representations and training objectives together, rather than treating encoder or decoder architecture as a sufficient explanation.

Information gathering places another demand on the decision process. CLM ranks separately encoded states and actions by their similarity~\citep{contrastivelm2026clm}. Mouselab instead requires evaluating whether an additional observation could improve a later choice. CLM stops immediately in every episode, as do Laya, Intern-Decision, and Minitaur. One possible explanation is that these models do not assign sufficient value to information acquisition under the evaluated protocol. Because this behavior occurs across different designs, it is not specific to one architecture. Without a control that varies the position or wording of the STOP option, a response bias toward stopping is not excluded.

The strongest evidence separating model design from decision execution comes from the backbone comparison and readout controls. Intern-Decision is reliably better than Qwen on 19 of 26 performance targets while retaining its backbone architecture, directing attention to fine-tuning for structured decisions and the evaluation pipeline. Conversely, changing the readout while holding model weights fixed reverses some rankings. Cognitive performance therefore depends on both the information represented by a model and how its scores are converted into actions.

These results motivate evaluating decision models through cognitive profiles alongside aggregate task scores. \bench{} provides a common set of tasks for examining how choices respond to feedback, memory demands, and opportunities to acquire information. For model developers, these profiles provide behavioral targets for testing changes to training objectives and history representations. Readout controls help establish how action selection contributes to performance while model weights remain fixed. The documented protocols and reference results also support comparisons across new checkpoints. Researchers can therefore examine whether an improvement extends across cognitive demands, which weaknesses remain, and how these patterns change as decision models develop.

\section{Conclusion}
\label{sec:conclusion}
\bench{} tests the cognitive strengths and limitations of five System One decision models, alongside three open comparison models, across 24 established tasks. Performance varies widely across these models. In them, strong autonomous performance does not guarantee accurate human-response prediction, and aggregate scores hide differences in responses to task demands. Extended free-text protocols are much slower than direct decisions in our measurements, while matched one-token generation adds little overhead. These cognitive tests complement accuracy and latency by checking whether fast, well-formed decisions use relevant evidence before software acts on them.

\section*{Limitations}
Coverage is limited to 24 tasks and five System One decision models. Adaptations include text conversion, supplied histories, Hybrid posterior summaries and free observation in autonomous Mouselab. Architecture, capacity, training data and input retention vary together, leaving architectural explanations unresolved. Minitaur's fine-tuning data share source experiments with some tasks (Section~\ref{sec:protocol}), and overlap for other models remains possible. API models differ in size, training, reasoning settings and infrastructure.

\section*{Ethical considerations}
The study uses only published human data and environments under their original terms and recruits no new participants. Restricted participant records are not redistributed, and model-generated behavior is kept separate from human observations.

\bibliographystyle{paperonecol}
\bibliography{main}

\begin{thebibliography}{27}
\providecommand{\natexlab}[1]{#1}
\providecommand{\url}[1]{\texttt{#1}}
\expandafter\ifx\csname urlstyle\endcsname\relax
  \providecommand{\doi}[1]{doi: #1}\else
  \providecommand{\doi}{doi: \begingroup \urlstyle{rm}\Url}\fi

\bibitem[Binz(n.d.)]{binzMinitaur}
Marcel Binz.
\newblock {Llama-3.1-Minitaur-8B}: Model card, n.d.
\newblock URL \url{https://huggingface.co/marcelbinz/Llama-3.1-Minitaur-8B}.
\newblock Hugging Face. Accessed 28 October 2026.

\bibitem[Binz et~al.(2025)Binz, Akata, Bethge, Br{\"a}ndle, Callaway, Coda-Forno, Dayan, Demircan, Eckstein, {\'E}ltet{\H{o}}, Griffiths, Haridi, Jagadish, Ji-An, Kipnis, Kumar, Ludwig, Mathony, Mattar, Modirshanechi, Nath, Peterson, Rmus, Russek, Saanum, Schubert, Schulze~Buschoff, Singhi, Sui, Thalmann, Theis, Truong, Udandarao, Voudouris, Wilson, Witte, Wu, Wulff, Xiong, and Schulz]{binz2025centaur}
Marcel Binz, Elif Akata, Matthias Bethge, Franziska Br{\"a}ndle, Fred Callaway, Julian Coda-Forno, Peter Dayan, Can Demircan, Maria~K. Eckstein, No{\'e}mi {\'E}ltet{\H{o}}, Thomas~L. Griffiths, Susanne Haridi, Akshay~K. Jagadish, Li~Ji-An, Alexander Kipnis, Sreejan Kumar, Tobias Ludwig, Marvin Mathony, Marcelo Mattar, Alireza Modirshanechi, Surabhi~S. Nath, Joshua~C. Peterson, Milena Rmus, Evan~M. Russek, Tankred Saanum, Johannes~A. Schubert, Luca~M. Schulze~Buschoff, Nishad Singhi, Xin Sui, Mirko Thalmann, Fabian~J. Theis, Vuong Truong, Vishaal Udandarao, Konstantinos Voudouris, Robert Wilson, Kristin Witte, Shuchen Wu, Dirk~U. Wulff, Huadong Xiong, and Eric Schulz.
\newblock A foundation model to predict and capture human cognition.
\newblock \emph{Nature}, 644\penalty0 (8078):\penalty0 1002--1009, 2025.
\newblock \doi{10.1038/s41586-025-09215-4}.
\newblock URL \url{https://doi.org/10.1038/s41586-025-09215-4}.

\bibitem[Callaway et~al.(2022)Callaway, van Opheusden, Gul, Das, Krueger, Griffiths, and Lieder]{callaway2022planning}
Frederick Callaway, Bas van Opheusden, Sayan Gul, Priyam Das, Paul~M. Krueger, Thomas~L. Griffiths, and Falk Lieder.
\newblock Rational use of cognitive resources in human planning.
\newblock \emph{Nature Human Behaviour}, 6\penalty0 (8):\penalty0 1112--1125, 2022.
\newblock \doi{10.1038/s41562-022-01332-8}.
\newblock URL \url{https://doi.org/10.1038/s41562-022-01332-8}.

\bibitem[Chen et~al.(2026)Chen, Liu, An, Wang, and Zhou]{chen2026system1arena}
Tianyu Chen, Xinyang Liu, Ruyi An, Zhendong Wang, and Mingyuan Zhou.
\newblock {System1-Arena: Which Decisions Can Be Left to System 1?}, 2026.
\newblock URL \url{https://arxiv.org/abs/2610.21244v1}.
\newblock Version 1, submitted 15 October 2026.

\bibitem[Coda-Forno et~al.(2024)Coda-Forno, Binz, Wang, and Schulz]{codaforno2024cogbench}
Julian Coda-Forno, Marcel Binz, Jane~X Wang, and Eric Schulz.
\newblock {CogBench}: a large language model walks into a psychology lab.
\newblock In \emph{Proceedings of the 41st International Conference on Machine Learning}, volume 235 of \emph{Proceedings of Machine Learning Research}, pp.\  9076--9108. PMLR, 2024.
\newblock URL \url{https://proceedings.mlr.press/v235/coda-forno24a.html}.

\bibitem[Collins \& Frank(2012)Collins and Frank]{collins2012rlwm}
Anne G.~E. Collins and Michael~J. Frank.
\newblock How much of reinforcement learning is working memory, not reinforcement learning? a behavioral, computational, and neurogenetic analysis.
\newblock \emph{European Journal of Neuroscience}, 35\penalty0 (7):\penalty0 1024--1035, 2012.
\newblock \doi{10.1111/j.1460-9568.2011.07980.x}.
\newblock URL \url{https://doi.org/10.1111/j.1460-9568.2011.07980.x}.

\bibitem[{Contrastive-LM}(2026)]{contrastivelm2026clm}
{Contrastive-LM}.
\newblock {CLM-v0.1-8B}: Model card, 2026.
\newblock URL \url{https://huggingface.co/Contrastive-LM/CLM-v0.1-8B}.
\newblock Hugging Face, revision e939398. Accessed 4 November 2026.

\bibitem[{Convai Innovations}(n.d.)]{convaiLaya}
{Convai Innovations}.
\newblock {Laya}: Official model card.
\newblock Hugging Face, n.d.
\newblock URL \url{https://huggingface.co/convaiinnovations/laya}.
\newblock Accessed 17 October 2026.

\bibitem[Daw et~al.(2011)Daw, Gershman, Seymour, Dayan, and Dolan]{daw2011model}
Nathaniel~D. Daw, Samuel~J. Gershman, Ben Seymour, Peter Dayan, and Raymond~J. Dolan.
\newblock Model-based influences on humans' choices and striatal prediction errors.
\newblock \emph{Neuron}, 69\penalty0 (6):\penalty0 1204--1215, 2011.
\newblock \doi{10.1016/j.neuron.2011.02.027}.

\bibitem[Grant \& Berg(1948)Grant and Berg]{grant1948cardsorting}
David~A. Grant and E.~A. Berg.
\newblock A behavioral analysis of degree of reinforcement and ease of shifting to new responses in a {Weigl}-type card-sorting problem.
\newblock \emph{Journal of Experimental Psychology}, 38\penalty0 (4):\penalty0 404--411, 1948.
\newblock \doi{10.1037/h0059831}.
\newblock URL \url{https://doi.org/10.1037/h0059831}.

\bibitem[Grattafiori et~al.(2024)Grattafiori, Dubey, Jauhri, Pandey, Kadian, Al-Dahle, et~al.]{grattafiori2024llama3}
Aaron Grattafiori, Abhimanyu Dubey, Abhinav Jauhri, Abhinav Pandey, Abhishek Kadian, Ahmad Al-Dahle, et~al.
\newblock The {Llama} 3 herd of models, 2024.
\newblock URL \url{https://arxiv.org/abs/2407.21783}.

\bibitem[Hou et~al.(2026{\natexlab{a}})Hou, Jiang, Li, Li, Lin, and Yamada]{hou2026wmip}
Dengzhe Hou, Lingyu Jiang, Deng Li, Zirui Li, Fangzhou Lin, and Kazunori~D Yamada.
\newblock {WMIP}: A working-memory-inspired programmable probe of cumulative state tracking in large language models, 2026{\natexlab{a}}.
\newblock URL \url{https://arxiv.org/abs/2603.27343v3}.
\newblock Version 3.

\bibitem[Hou et~al.(2026{\natexlab{b}})Hou, Jiang, Lin, and Yamada]{hou2026cogbattery}
Dengzhe Hou, Lingyu Jiang, Fangzhou Lin, and Kazunori~D Yamada.
\newblock {CogBattery}: Human-referenced evaluation of cognitive task behavior and structure in language models, 2026{\natexlab{b}}.
\newblock URL \url{https://arxiv.org/abs/2607.24999v2}.
\newblock Version 2.

\bibitem[Jones et~al.(2024)Jones, Trott, and Bergen]{jones2024epitome}
Cameron~R. Jones, Sean Trott, and Benjamin Bergen.
\newblock Comparing humans and large language models on an experimental protocol inventory for theory of mind evaluation ({EPITOME}).
\newblock \emph{Transactions of the Association for Computational Linguistics}, 12:\penalty0 803--819, 2024.
\newblock \doi{10.1162/tacl_a_00674}.
\newblock URL \url{https://aclanthology.org/2024.tacl-1.45/}.

\bibitem[Kolluri et~al.(2025)Kolluri, Wu, Park, and Bernstein]{kolluri2025socsci210}
Akaash Kolluri, Shengguang Wu, Joon~Sung Park, and Michael~S. Bernstein.
\newblock Finetuning {LLM}s for human behavior prediction in social science experiments.
\newblock In Christos Christodoulopoulos, Tanmoy Chakraborty, Carolyn Rose, and Violet Peng (eds.), \emph{Proceedings of the 2025 Conference on Empirical Methods in Natural Language Processing}, pp.\  30096--30111, Suzhou, China, November 2025. Association for Computational Linguistics.
\newblock ISBN 979-8-89176-332-6.
\newblock \doi{10.18653/v1/2025.emnlp-main.1530}.
\newblock URL \url{https://aclanthology.org/2025.emnlp-main.1530/}.

\bibitem[Namazova et~al.(2025)Namazova, Brondetta, Strittmatter, Nassar, and Musslick]{namazova2025centaur}
Sabrina Namazova, Alessandra Brondetta, Younes Strittmatter, Matthew Nassar, and Sebastian Musslick.
\newblock {Not Yet AlphaFold for the Mind: Evaluating Centaur as a Synthetic Participant}, 2025.
\newblock URL \url{https://arxiv.org/abs/2508.07887v1}.
\newblock Version 1.

\bibitem[Nie et~al.(2025)Nie, Su, Chang, Lee, Chi, Le, and Chen]{nie2025evolve}
Allen Nie, Yi~Su, Bo~Chang, Jonathan Lee, Ed~H. Chi, Quoc~V Le, and Minmin Chen.
\newblock {EVOLvE}: Evaluating and optimizing {LLMs} for in-context exploration.
\newblock In \emph{Proceedings of the 42nd International Conference on Machine Learning}, volume 267 of \emph{Proceedings of Machine Learning Research}, pp.\  46346--46376. PMLR, 2025.
\newblock URL \url{https://proceedings.mlr.press/v267/nie25b.html}.

\bibitem[{Qwen Team}(n.d.{\natexlab{a}})]{qwen35_4b}
{Qwen Team}.
\newblock {Qwen3.5-4B}: Model card, n.d.{\natexlab{a}}.
\newblock URL \url{https://huggingface.co/Qwen/Qwen3.5-4B}.
\newblock Hugging Face. Accessed 28 October 2026.

\bibitem[{Qwen Team}(n.d.{\natexlab{b}})]{qwen3_8b}
{Qwen Team}.
\newblock {Qwen3-8B}: Model card, n.d.{\natexlab{b}}.
\newblock URL \url{https://huggingface.co/Qwen/Qwen3-8B}.
\newblock Hugging Face. Accessed 4 November 2026.

\bibitem[Sharma et~al.(2026)Sharma, Duncan, Marceau~Caron, and Vollmer]{sharma2026typedbench}
Rahul Sharma, Andrew~B. Duncan, Ga\'{e}tan Marceau~Caron, and Sebastian~J. Vollmer.
\newblock {TypedBench: A Benchmark for Calibration, Framing Sensitivity, and Cost in System One Decision Models}, 2026.
\newblock URL \url{https://arxiv.org/abs/2610.11392v2}.
\newblock Version 2, revised 9 October 2026.

\bibitem[Sun et~al.(2026)Sun, Shleifer, and Belinkov]{sun2026choose}
Johnathan Sun, Andrei Shleifer, and Yonatan Belinkov.
\newblock Do {LLM}s choose like humans? using cognitive theory to evaluate {LLM} decision-making.
\newblock In Hung-Yi Lee, Sunipa Dev, Jessy Li, and Desmond Elliott (eds.), \emph{Findings of the {A}ssociation for {C}omputational {L}inguistics: {EMNLP} 2026}, pp.\  46552--46569, Budapest, Hungary, October 2026. Association for Computational Linguistics.
\newblock ISBN 979-8-89176-462-0.
\newblock URL \url{https://aclanthology.org/2026.findings-emnlp.2177/}.

\bibitem[Tong et~al.(2026)Tong, Yue, Zhao, Lin, Jia, Chen, Sun, Zhang, and Zeng]{tong2026cogtom}
Haibo Tong, Zeyang Yue, Feifei Zhao, Erliang Lin, Lu~Jia, Ruolin Chen, Yinqian Sun, Qian Zhang, and Yi~Zeng.
\newblock {C}og{T}o{M}: A comprehensive theory of mind benchmark inspired by human cognition for large language models.
\newblock In Maria Liakata, Viviane~P. Moreira, Jiajun Zhang, and David Jurgens (eds.), \emph{Proceedings of the 64th Annual Meeting of the {A}ssociation for {C}omputational {L}inguistics (Volume 1: Long Papers)}, pp.\  31383--31425, San Diego, California, United States, July 2026. Association for Computational Linguistics.
\newblock ISBN 979-8-89176-390-6.
\newblock \doi{10.18653/v1/2026.acl-long.1448}.
\newblock URL \url{https://aclanthology.org/2026.acl-long.1448/}.

\bibitem[{TypeSafe}(n.d.)]{typesafeJev}
{TypeSafe}.
\newblock {System One}.
\newblock Official documentation, n.d.
\newblock URL \url{https://docs.typesafe.ai/concepts/system-one}.
\newblock Accessed 17 October 2026.

\bibitem[{vLLM Semantic Router}(2026)]{vllmsr2026nox}
{vLLM Semantic Router}.
\newblock {Decision-2.0-Nox-4B}: Model card, 2026.
\newblock URL \url{https://huggingface.co/vllm-sr/Decision-2.0-Nox-4B}.
\newblock Hugging Face, revision ae269ec. Accessed 4 November 2026.

\bibitem[Wilson et~al.(2014)Wilson, Geana, White, Ludvig, and Cohen]{wilson2014exploration}
Robert~C. Wilson, Andra Geana, John~M. White, Elliot~A. Ludvig, and Jonathan~D. Cohen.
\newblock Humans use directed and random exploration to solve the explore--exploit dilemma.
\newblock \emph{Journal of Experimental Psychology: General}, 143\penalty0 (6):\penalty0 2074--2081, 2014.
\newblock \doi{10.1037/a0038199}.
\newblock URL \url{https://doi.org/10.1037/a0038199}.

\bibitem[Ying et~al.(2026)Ying, Wu, Wang, Aarya, Schulze~Buschoff, Chen, Collins, de~Varda, Fu, Houlihan, Jagadish, Jiang, Kiegeland, Kurumisawa, Liu, Liu, Ma, McGregor, Strittmatter, Tsvilodub, Vigly, Wu, Xu, Yun, Allen, Brooke-Wilson, Christian, Fedorenko, Frank, Franke, Gao, Gershman, Hawkins, Hu, Jara-Ettinger, Kleiman-Weiner, Levine, Linzen, Lu, O'Donnell, Ong, Piantadosi, Saxe, Schulz, Shu, Sosa, Sucholutsky, Zhi-Xuan, Ullman, Xu, Yildirim, Zhu, Griffiths, Gerstenberg, Smith, and Tenenbaum]{ying2026coggym}
Lance Ying, Jinzhou Wu, Yingshan~Susan Wang, Shivam Aarya, Luca~M. Schulze~Buschoff, Harry Chen, Katherine~M. Collins, Andrea de~Varda, Shuhao Fu, Sean~Dae Houlihan, Akshay~K. Jagadish, Guangyuan Jiang, Samuel Kiegeland, Tetsu Kurumisawa, Rongzhi Liu, Ryan Liu, Ningshan Ma, Kathryn McGregor, Younes Strittmatter, Polina Tsvilodub, Jacob~Hoover Vigly, Sarah Wu, Enjie Xu, Yiling Yun, Kelsey Allen, Tyler Brooke-Wilson, Brian Christian, Evelina Fedorenko, Michael~C. Frank, Michael Franke, Tao Gao, Samuel~J. Gershman, Robert~D. Hawkins, Jennifer Hu, Julian Jara-Ettinger, Max Kleiman-Weiner, Sydney Levine, Tal Linzen, Hongjing Lu, Timothy O'Donnell, Desmond~C. Ong, Steven~T. Piantadosi, Rebecca Saxe, Eric Schulz, Tianmin Shu, Felix~A. Sosa, Ilia Sucholutsky, Tan Zhi-Xuan, Tomer Ullman, Fei Xu, Ilker Yildirim, Jian-Qiao Zhu, Thomas~L. Griffiths, Tobias Gerstenberg, Kevin Smith, and Joshua~B. Tenenbaum.
\newblock {CogGym: Towards Large-Scale Comparative Evaluation of Human and Machine Cognition}, 2026.
\newblock URL \url{https://arxiv.org/abs/2609.21259v1}.
\newblock Version 1.

\bibitem[Zhu et~al.(2026)Zhu, Xu, He, Chai, Hu, Zheng, Luo, Li, Zhang, and Cheng]{internlm2026decision}
Tong Zhu, Zhe Xu, Zefeng He, Yunkai Chai, Xuyang Hu, Dehong Zheng, Yun Luo, Yafu Li, Qi~Zhang, and Yu~Cheng.
\newblock {Intern-Decision}: Efficient multimodal decision making with autoregressive masked language modeling, 2026.
\newblock URL \url{https://arxiv.org/abs/2610.22131}.

\end{thebibliography}
\appendix
\renewcommand{\paragraph}[1]{\par\noindent\phantomsection\textbf{#1}\quad}
\raggedbottom
\section{Protocol and reference details}
\label{app:protocol}\label{app:details}\label{app:selected-api-extension}

\paragraph{Readouts.}
Laya uses the English checkpoint at \texttt{convaiinnovations/laya}.
Horizon uses SDK probability sampling for Laya, the source sampled-token rule for Llama and Minitaur, and native legal argmax for Intern, Qwen, Nox, and CLM. The source rule samples one token from the full vocabulary at temperature 1 with top-p 0.9 and top-k 50. This truncation can remove the less likely key, which the legal sampling of Table~\ref{tab:p3-final-horizon} never does. WCST uses legal argmax for Laya, Intern, Qwen, Nox, and CLM. Llama and Minitaur use full-vocabulary next-token argmax for human prediction and sampled generation for autonomous execution. Legal argmax and categorical sampling are compared across all seven open models in both tasks.
Minitaur's nonliteral reports use the number before its closing marker, because its generation continues past that marker.

\paragraph{Context and history.}
For multiple-cue judgment, Llama, Minitaur, Nox, CLM, and Jev use full prefixes and the other models use compact history. Nox uses its own runtime with a 16,384-token context, so no input is truncated. \pagebreak[4]
CLM uses the official vLLM pooling server for Qwen3-8B. That setup keeps the first 2,048 tokens of each text, so in 11 modes it cuts the end of long states, including the question, on some calls, such as 66\% of the original WCST calls. A control that keeps the last 2,048 tokens instead changes the scores but leaves every comparison with uniform choice unchanged. Under its original native readout, CLM always chooses machine H in autonomous Horizon and always stops in autonomous Mouselab, so it scores exactly 50\% and zero reward there.

\begin{table}[!htbp]
\centering\footnotesize
\setlength{\tabcolsep}{4pt}
\begin{tabular}{@{}lrrrrrr@{}}
\toprule
 & \multicolumn{3}{c}{Legal argmax} & \multicolumn{3}{c}{Legal sampling} \\
\cmidrule(lr){2-4}\cmidrule(lr){5-7}
Model & Accuracy $\uparrow$ & Persev. $\downarrow$ & Set-loss $\downarrow$ & Accuracy $\uparrow$ & Persev. $\downarrow$ & Set-loss $\downarrow$ \\
\midrule
\rowcolor{tablegray} Uniform & 25.00 & 17.13 & 17.13 & 25.00 & 17.13 & 17.13 \\
Laya & 25.47 & 15.24 & 18.16 & 25.18 & 16.61 & 17.19 \\
Intern & 27.90 & 21.25 & 15.84 & 26.32 & 17.00 & 16.90 \\
Llama & \underline{32.37} & 35.49 & 10.71 & \textbf{31.01} & 23.25 & 15.64 \\
Minitaur & \textbf{34.28} & 48.46 & 3.71 & 27.07 & 16.91 & 16.83 \\
Qwen & 25.20 & 15.88 & 17.88 & 25.76 & 16.37 & 17.98 \\
Nox & 32.14 & 32.53 & 11.88 & \underline{28.70} & 19.19 & 16.81 \\
CLM & 25.71 & 14.96 & 18.08 & 25.81 & 16.04 & 17.82 \\
\bottomrule
\end{tabular}
\caption{WCST readouts over 75 held-out participants (12,754 choices per model and readout). Values are participant-mean percentages. Bold and underlining mark the best and second-best open-model accuracies before rounding. Uniform is repeated as the gray reference. Error definitions follow Table~\ref{tab:wcst}.}
\label{tab:wcst-readouts}
\end{table}
\FloatBarrier

\paragraph{API models.}
Jev 1.14 executes its native choice, and its two-decimal probabilities are normalized without a floor. The Claude5.5 models use adaptive thinking, medium effort, and a 4,096-token cap and answer on a final line of text, and an invalid answer counts as wrong and ends a closed-loop run. Haiku\textquotesingle{}s NLL uses probabilities it states in JSON, a different readout from the open models. Several stated politeness probabilities closely match the published human proportions, so exposure to these public data cannot be excluded.

\paragraph{Reference provenance.}
\label{app:reference-sources}\label{app:new-traditional-references}\label{app:remaining-references}\label{app:recovered-references}
Across the 27 modes, references comprise 16 participant fits, 5 summaries of other participants' responses, 4 policies without human fitting, and 2 analytic solutions or answer keys. The Horizon RW reference uses 25 training and 6 held-out participants, and the WCST SL reference uses 300 training and 75 held-out participants. RLWM replays archived fits for the same participants being evaluated, so it is not a held-out comparison. The complete catalogue of fitting data, splits, and implementations is provided in the project repository. In recent probes, the prompt shows the current letter set, so perfect reference accuracy does not establish a working-memory mechanism. Stationary bandits use the public FullBench conditions rather than the original publication's protocol.

\FloatBarrier
\section{Statistical support for the main results}
\label{app:tasks}\label{app:construction}\label{app:accepted-controls}\label{app:table2-controls}
\par\noindent
\begin{minipage}{\linewidth}
\centering
\scriptsize
\setlength{\tabcolsep}{3pt}
\makeatletter\def\@captype{table}\makeatother
\caption{RLWM accuracy and condition contrasts under complete human-block histories. Gray rows denote the Human and Uniform references.}
\label{tab:rlwm-models}
\vspace{3pt}
\begin{tabular}{@{}lrrr@{}}
\toprule
 & Overall & Size 6 minus 2 & Pres. 10 minus 1 \\
Source & (\%, $n=79$) & (pp, $n=79$) & size 6 (pp, $n=69$) \\
\midrule
\rowcolor{tablegray} Human & 80.24 & $-$12.74 [$-$14.54, $-$10.99] & +46.35 [41.13, 51.53] \\
\rowcolor{tablegray} Uniform & 33.33 & +0.00 & +0.00 \\
Laya & 34.85 & $-$7.17 [$-$8.53, $-$5.79] & +0.21 [$-$2.46, 2.69] \\
Intern & \underline{77.83} & $-$11.37 [$-$12.80, $-$9.96] & +61.49 [58.09, 64.84] \\
Llama & 41.78 & $-$16.73 [$-$17.94, $-$15.46] & +10.49 [7.27, 13.62] \\
Minitaur & 49.09 & $-$11.34 [$-$12.58, $-$10.12] & +11.10 [7.34, 15.11] \\
Qwen & 59.39 & $-$20.34 [$-$22.21, $-$18.41] & +19.75 [15.71, 23.87] \\
Nox & \textbf{86.30} & $-$4.84 [$-$5.87, $-$3.79] & +58.65 [56.32, 60.99] \\
CLM & 36.74 & +0.99 [$-$0.31, 2.29] & +2.03 [$-$1.77, 5.71] \\
\bottomrule
\end{tabular}
\end{minipage}
\par

\newpage

Table~\ref{tab:rlwm-models} reports RLWM accuracy over all 52,911 eligible responses from 79 participants given complete human-block histories. Scores use hidden correct actions and equal participant weights (Table~\ref{tab:task-summary} weights responses). Pres. means presentation. Contrasts pair the participants listed. Brackets show pointwise 95\% participant-bootstrap intervals (5,000 resamples, uncorrected). Bold and underlining mark the best and second-best overall open-model accuracies.

Table~\ref{tab:uniform-intervals} reports 95\% percentile bootstrap intervals of each model's difference from Uniform on the 26 directional targets (10,000 resamples of the listed units). Intervals are unadjusted for multiple comparisons. Weather prediction has 2 participants and multi-attribute choice and Horizon prediction have 6 each, so their intervals reflect these small samples.

$+$ and $-$ mark intervals wholly better or worse than Uniform, and $\circ$ marks intervals containing zero. $+\infty$ denotes an infinite score in every resample. \textsuperscript{c} marks constant unit scores, reported only by the sign of the point difference ($0$ for zero).

\begingroup
\footnotesize\setlength{\tabcolsep}{4pt}
\setlength{\LTcapwidth}{\linewidth}
\setlength{\LTpre}{6pt}
\begin{longtable}{@{}ll cccccccc@{}}
\caption{Each model against Uniform on the 26 directional targets of Table~\ref{tab:task-summary}. The gray row gives counts, with the highest and second highest among open models marked in bold and underlined.}
\label{tab:uniform-intervals}\\
\toprule
Target & Resampled units & Laya & Intern & Llama & Minitaur & Qwen & Nox & CLM & Jev \\
\midrule
\endfirsthead
\multicolumn{10}{l}{Table \thetable\ (continued)} \\
\toprule
Target & Resampled units & Laya & Intern & Llama & Minitaur & Qwen & Nox & CLM & Jev \\
\midrule
\endhead
\midrule
\multicolumn{10}{r}{Continued on next page} \\
\endfoot
\endlastfoot
Bayesian urn & 200 generated runs & $+$ & $+$ & $+$ & $+$ & $+$ & $+$ & $+$ & $+$ \\
Multiple-cue judgment & 28 participants & $-$ & $+$ & $-$ & $+$ & $-$ & $+$ & $-$ & $+\infty$ \\
Multi-attribute choice & 6 participants & $\circ$ & $+$ & $-$ & $\circ$ & $\circ$ & $+$ & $-$ & $+$ \\
Probability coherence & 15 participants & $\circ$ & $+$ & $+$ & $+$ & $+$ & $+$ & $-$ & $+$ \\
SHJ categories & 10 participants & $\circ$ & $+$ & $+$ & $+$ & $+$ & $+$ & $\circ$ & $+$ \\
Medin categories & 42 participants & $-$ & $+$ & $+$ & $+$ & $+$ & $+$ & $+$ & $+$ \\
Weather prediction & 2 participants & $+$ & $+$ & $-$ & $+$ & $+$\textsuperscript{c} & $+$ & $+$ & $+$ \\
RLWM & 79 participants & $+$ & $+$ & $+$ & $+$ & $+$ & $+$ & $+$ & $+$ \\
Recent probes & 51 participants & $\circ$ & $+$ & $+$ & $-$ & $+$ & $+$ & $+$ & $+$ \\
WCST, prediction & 75 participants & $-$ & $\circ$ & $-$ & $+$ & $-$ & $-$ & $-$ & $-$ \\
WCST, autonomous & 75 participants & $\circ$ & $+$ & $+$ & $+$ & $\circ$ & $+$ & $+$ & $+$ \\
Restless bandit & 10 runs & $+$ & $+$ & $+$ & $-$ & $\circ$ & $+$ & $\circ$ & $+$ \\
Instrumental learning & 10 runs & $\circ$ & $\circ$ & $\circ$ & $+$ & $\circ$ & $+$ & $\circ$ & $+$ \\
Horizon, prediction & 6 participants & $\circ$ & $+$ & $+$ & $+$ & $\circ$ & $+$ & $-$ & $+$ \\
Horizon, autonomous & 31 participant timelines & $+$ & $+$ & $+$ & $+$ & $+$ & $+$ & $0$\textsuperscript{c} & $+$ \\
Stationary bandits & 480 episodes within condition & $+$ & $+$ & $\circ$ & $-$ & $+$ & $+$ & $+$ & $+$ \\
Hybrid sampling & 20 participants & $-$ & $+$ & $-$ & $+$ & $-$ & $-$ & $-$ & $+\infty$ \\
Mouselab, prediction & 95 participants & $-$ & $-$ & $-$ & $-$ & $-$ & $+$ & $-$ & $+$ \\
Mouselab, autonomous & 95 schedule slots & $-$\textsuperscript{c} & $-$\textsuperscript{c} & $-$ & $-$\textsuperscript{c} & $+$ & $-$ & $-$\textsuperscript{c} & $+$ \\
BART & 20 runs & $-$ & $\circ$ & $-$ & $-$\textsuperscript{c} & $-$\textsuperscript{c} & $+$ & $+$ & $+$ \\
Two-step & 100 runs & $\circ$ & $+$ & $\circ$ & $\circ$ & $\circ$ & $+$ & $\circ$ & $+$ \\
False-belief QA & 999 story blocks & $-$ & $+$ & $+$ & $+$ & $+$ & $+$ & $+$ & $+$ \\
Politeness & 80 items & $-$ & $+$ & $-$ & $\circ$ & $\circ$ & $+$ & $-$ & $+\infty$ \\
Pragmatic interpretation & 170 items & $-$ & $+$ & $-$ & $+$ & $\circ$ & $+$ & $+$ & $+\infty$ \\
Nonliteral number & 60 items & $+$ & $+$ & $+$ & $+$ & $+$ & $+$ & $+$ & $+$ \\
Self-assessment & 3866 participants & $+\infty$ & $+$ & $-$ & $+$ & $-$ & $+$ & $-$ & $+\infty$ \\
\midrule
\rowcolor{tablegray} Reliably better than Uniform & & 7 & \underline{21} & 12 & 17 & 11 & \textbf{23} & 11 & 20 \\
\bottomrule
\end{longtable}
\endgroup

\FloatBarrier

\section{Scoring and timing definitions}
\label{app:scoring-reproduction}

\paragraph{Probability being scored.}
The reported scores use complete task modes or complete API targets. The human-response NLL entries in Table~\ref{tab:task-summary} score probabilities over the task's legal response support. For a causal decoder with answer-token logits $z_{k(a)}$, the legal distribution is
\[
q(a\mid x,\mathcal A)=\frac{\exp z_{k(a)}}{\sum_{b\in\mathcal A}\exp z_{k(b)}}.
\]
The token index $k(a)$ is checked after the complete prompt, and no multi-token answer is scored as its first token. Laya's SDK probabilities are rounded to four decimals and renormalized over the legal keys. Nox and Jev normalize their returned legal probabilities without a floor, so Jev's two-decimal rounding can give exact zeros. CLM supplies a softmax over its scaled cosine scores, and Haiku's NLL uses the distribution it states in JSON. Full-vocabulary diagnostics, native actions, and sampled actions are recorded separately and never replace these entries. Table~\ref{tab:nll-definitions} gives each target's support, zero handling, and aggregation.

\begin{table}[!htp]
\centering\footnotesize
\begin{tabularx}{\linewidth}{@{}lLL@{}}
\toprule
Prediction target & Support and primary score & Aggregation \\
\midrule
Multiple-cue judgment & Original nine response levels, $-\log q(a_{\rm human})$, no epsilon & Response-weighted mean \\
Multi-attribute choice & Two legal product keys, $-\log q(a_{\rm human})$, no epsilon & Response-weighted mean \\
WCST & Four card keys, $-\log(q(a_{\rm human})+10^{-12})$ & Mean of participant means \\
Horizon & H/I keys, $-\log(q(a_{\rm human})+10^{-8})$ & Response-weighted mean \\
Hybrid sampling & Four source action keys, $-\log q(a_{\rm human})$, no epsilon & Pooled held-out actions \\
Mouselab & Currently legal reveals and STOP, $-\log q(a_{\rm human})$, no epsilon & Mean over scored decisions \\
Politeness & Each item's legal answers, cross-entropy against its normalized human counts, no epsilon & Weighted by human response counts \\
Pragmatic interpretation & Each item's legal answers, cross-entropy against its normalized human counts, no epsilon & Weighted by human response counts \\
Self-assessment & Original legal response labels for each item, $-\log q(a_{\rm human})$, no epsilon & Response-weighted mean \\
\bottomrule
\end{tabularx}
\caption{The probability support, zero handling, and aggregation of the nine primary human-prediction targets. Source weights and folds are retained. For cross-entropy, a zero receives infinite loss only when the corresponding human count is positive.}
\label{tab:nll-definitions}
\end{table}

\paragraph{Scoring conventions.}
The source scorers and their denominators are unchanged, so an invalid executed action counts as an error, and no smoothing makes an infinite primary score finite. The epsilons in Table~\ref{tab:nll-definitions} belong only to the source WCST and Horizon scorers. Uniform uses source weights and tie credit over legal actions or reports, with exact expectations for autonomous modes. Temporal discounting covers 6 realized-reward questions for Laya, Intern, Llama, Nox, and CLM and 10 for Minitaur and Qwen, excludes payment items, and has Uniform at 50\% only on these questions. False-belief QA uses an answer key because it has no human responses.

\paragraph{Timing.}
The controlled timer brackets tokenization, adapter checks, model forward, and readout with CUDA synchronization, and it excludes model loading and the 20 warm-up calls per mode. For Llama and Qwen, we compare a legal-choice forward pass with greedy one-token generation on 200 paired inputs in each of six modes. Checkpoints, prompts, BF16 precision, batch size, GPU, and cache settings are matched. Median request-level generation/forward latency ratios range from 1.0024 to 1.0202. All timings are retained, but unconstrained one-token outputs are not evaluated as complete answers. Full settings and per-mode results are in the project repository.

API latency includes network and provider service time, and the tested target sets differ. Jev's median is 0.35 seconds over all 27 modes, against 2.60 to 3.59 seconds for Claude5.5 with adaptive thinking. Recorded costs are US\$0.045 per 1,000 decisions for Jev and US\$0.34 to 8.96 for Claude (API timing table in the project repository). These measurements describe service protocols rather than isolate model speed.

\FloatBarrier

\section{Further related work}
\label{app:related}
CogToM builds a theory-of-mind benchmark from 46 paradigms \citep{tong2026cogtom}, and EPITOME compares five language models with human baselines on six theory-of-mind experiments \citep{jones2024epitome}. SocSci210 finetunes language models on 170 of 210 social science experiments to predict human response distributions \citep{kolluri2025socsci210}. \citet{sun2026choose} find that context shifts LLM product choices in human-like ways but does not reliably reweight attention between features. Horizon manipulates future opportunities and information asymmetry \citep{wilson2014exploration}, and BanditBench supports interactive choices with raw or summary histories and regret analysis \citep{nie2025evolve}. Mouselab studies information acquisition and its consequences \citep{callaway2022planning}.

\end{document}